\documentclass[letterpaper, 10 pt, conference]{ieeeconf}

\IEEEoverridecommandlockouts                              
\usepackage{graphics}
\usepackage{epsfig}
\usepackage{mathptmx} 
\usepackage{times}    
\usepackage{amsmath}
\usepackage{amssymb}
\usepackage{amsfonts}
\usepackage{algorithm}
\usepackage{algorithmicx} 
\usepackage{algpseudocode} 
\usepackage{multirow}
\usepackage{cite}
\usepackage{subfigure}
\usepackage{stfloats}
\usepackage{capt-of} 
\usepackage{textcomp}
\usepackage{array}
\usepackage{rotating}

\makeatletter
\let\NAT@parse\undefined
\makeatother
\usepackage{url}
\usepackage[table]{xcolor}
\usepackage{colortbl}
\usepackage{hyperref}
\hypersetup{colorlinks=true, linkcolor=red, urlcolor=blue, citecolor=green, anchorcolor=blue}

\makeatletter
\newcommand{\removelatexerror}{\let\@latex@error\@gobble}
\makeatother

\def\BibTeX{{\rm B\kern-.05em{\sc i\kern-.025em b}\kern-.08em
    T\kern-.1667em\lower.7ex\hbox{E}\kern-.125emX}}

\title{\LARGE \bf
CLASP: Closed-loop Asynchronous Spatial Perception for Open-vocabulary Desktop Object Grasping}

\author{
Yiran Ling$^{1,2,*}$ \quad Wenxuan Li$^{1,2,*}$ \quad Siying Dong$^{1,2,*}$ \quad Yize Zhang$^{1,2}$ \\
Xiaoyao Huang$^{4}$ \quad Jing Jiang$^{1,2}$ \quad Ruonan Li$^{3}$ \quad Jie Liu$^{1,2,\dag}$ \\
$^{1}$Harbin Institute of Technology \quad $^{2}$National Key Laboratory of Smart Farm Technologies and Systems \\
$^{3}$Peng Cheng Laboratory \quad $^{4}$Northeastern University \\
\texttt{jieliu@hit.edu.cn} \\
$^{*}$Equal contribution. \quad $\dag$Corresponding author.
}

\begin{document}


\maketitle
\thispagestyle{empty}
\pagestyle{empty}

\nocite{*}

\begin{abstract}
Robot grasping of desktop object is widely used in intelligent manufacturing, logistics, and agriculture.Although vision-language models (VLMs) show strong potential for robotic manipulation, their deployment in low-level grasping faces key challenges: scarce high-quality multimodal demonstrations, spatial hallucination caused by weak geometric grounding, and the fragility of open-loop execution in dynamic environments. To address these challenges, we propose Closed-Loop Asynchronous Spatial Perception(CLASP), a novel  asynchronous closed-loop framework that integrates multimodal perception, logical reasoning, and state-reflective feedback. First, we design a Dual-Pathway Hierarchical Perception module that decouples high-level semantic intent from geometric grounding. The design guides the output of the inference model and the definite action tuples, reducing spatial illusions. Second, an Asynchronous Closed-Loop Evaluator is implemented to compare pre- and post-execution states, providing text-based diagnostic feedback to establish a robust error-correction loop and improving the vulnerability of traditional open-loop execution in dynamic environments. Finally, we design a scalable multi-modal data engine that automatically synthesizes high-quality spatial annotations and reasoning templates from real and synthetic scenes without human teleoperation. Extensive experiments demonstrate that our approach significantly outperforms existing baselines, achieving an 87.0\% overall success rate. Notably, the proposed framework exhibits remarkable generalization across diverse objects, bridging the sim-to-real gap and providing exceptional robustness in geometrically challenging categories and cluttered scenarios.
\end{abstract}
\section{INTRODUCTION}


With the continuous integration of artificial intelligence and robotics, enabling robots to perform intelligent grasping according to natural language instructions has become a core requirement for autonomous manipulation. In open and unstructured environments, grasping under desktop-like planar constraints is particularly important because objects are diverse, layouts change frequently, and interaction often occurs in close proximity to people. Such capabilities are crucial across many practical scenarios, including retail settings like supermarkets, office desktops and shared workspaces, and other everyday service contexts where robots must retrieve, organize, or deliver items safely and efficiently. Grasping in real-world scenarios requires robots to not only perceive the geometric structure of objects and ensure grasping stability, but also understand complex semantic constraints. 

Existing robotic grasping approaches can be mainly categorized into two paradigms: rule-based grasping methods and recent Learning-based grasping methods. Rule-based robotic grasping methods \cite{b38,b10} typically rely on geometric information of objects and make insufficient use of the semantic information of the objects themselves. Based on learning methods such as vision-language models(VLM) and Visual-Language-Action(VLA) models~\cite{rt22023arxiv,black2024pi0,kimOpenVLAOpenSourceVisionLanguageAction2024}, it exhibits strong generalization ability for object manipulation. Although the Visual-Language-Action(VLA) models also perform well in grasping desktop objects, they are highly dependent on post-training. In recent years, the development of vision-language models (VLMs) \cite{b23,b24,b25} has introduced semantic understanding into grasping tasks. By incorporating affordance information into multimodal pre-training data, VLM can assist robots in grasping open vocabulary objects under natural language commands. Despite promising strides, deploying VLMs for low-level robotic manipulation faces critical bottlenecks:
1) a severe scarcity of high-quality multi-modal demonstration data;
2) VLMs' lack of fine-grained geometric grounding, leading to spatial hallucinations.

To address these challenges, we propose a novel closed-loop framework that bridges semantic reasoning with physical execution. By combining hierarchical perception with asynchronous closed-loop feedback, spatial illusions in robot grasping based on visual language models can be alleviated, bridging the semantic and geometric gap. The main contributions of this paper are summarized as follows:
\begin{itemize}
    \item We introduce a dual-pathway hierarchical perception approach that decouples semantic intent from geometric grounding. By outputting structured action tuples (\textit{e.g.}, 2D contact points and angles), this method mitigates spatial hallucinations and enables deterministic control.
    \item We design an asynchronous closed-loop evaluator. In this evaluator, a \textit{Judger} module compares pre- and post-execution states and generates diagnostic feedback. This establishes a robust error-correction mechanism for improved task success in dynamic environments.
    \item We develop a scalable multi-modal data engine. Within this engine, an automated pipeline synthesizes high-quality demonstration data (visual prompts, masks, grasp poses) without human teleoperation.
\end{itemize}

\section{RELATED WORKS}

\subsection{Basic Grasp Detection Models}

Existing grasping research mainly optimizes performance along a single dimension, lacking a unified framework that integrates geometric, semantic, and adaptive capabilities. For example, GraspNet\cite{b38} and AnyGrasp\cite{b10} directly regress 6-DoF grasps from point clouds or RGB-D images, achieving accurate geometric evaluation but ignoring semantic cues. \cite{b11,b14,b13} enhance stability via geometric modeling or tactile feedback, yet rely on complex designs or specialized hardware. Task-aware approaches \cite{b12,b16,b20} model gestures or manipulation relationships to capture intent, but require extra modalities and show limited cross-scenario generalization. Moreover, environment- or hardware-specific optimizations \cite{b15,b17,b18,b19} improve robustness in constrained settings, while remaining scenario- or task-dependent.

\subsection{Learning-based Grasping and Manipulation}

Learning-based grasping methods leverage large models for semantic understanding and reasoning in open-world scenarios. Early works \cite{b21,b22} adopt contrastive learning for multi-modal alignment but lack complex instruction reasoning. \cite{b23,b24,b25} introduce large-scale pre-training frameworks linking visual encoders and language models via projection layers, enabling instruction following and visual reasoning, yet remain unoptimized for robotic interaction. Building on this, several studies directly apply vision-language models (VLMs) to grasping. Methods in \cite{b27,b8} reformulate affordance recognition as VLM-based localization, enabling semantic-guided ROI generation but suffering from error accumulation in two-stage designs. \cite{b28} incorporates task-level reasoning for cluttered environments, improving strategic planning at the cost of real-time performance. End-to-end VLA models\cite{rt22023arxiv,black2024pi0}like OpenVLA~\cite{kimOpenVLAOpenSourceVisionLanguageAction2024}, directly input vision-language inputs into robot actions and achieve generalizable task execution capabilities, but require extensive fine-tuning data and large model scales, limiting their effectiveness.
deployment in long-tail and resource-constrained settings.Recent work increasingly explores dexterous manipulation for complex tasks, such as DexVLG \cite{DexVLG} and DexGrasp \cite{DexGrasp}. However, high hardware costs, limited compatibility with common configurations, and the lack of standardized datasets hinder practical deployment.

\section{Method}
\subsection{Problem Formulation}

\begin{figure*}[t]
  \centering
  \includegraphics[width=0.7\textwidth]{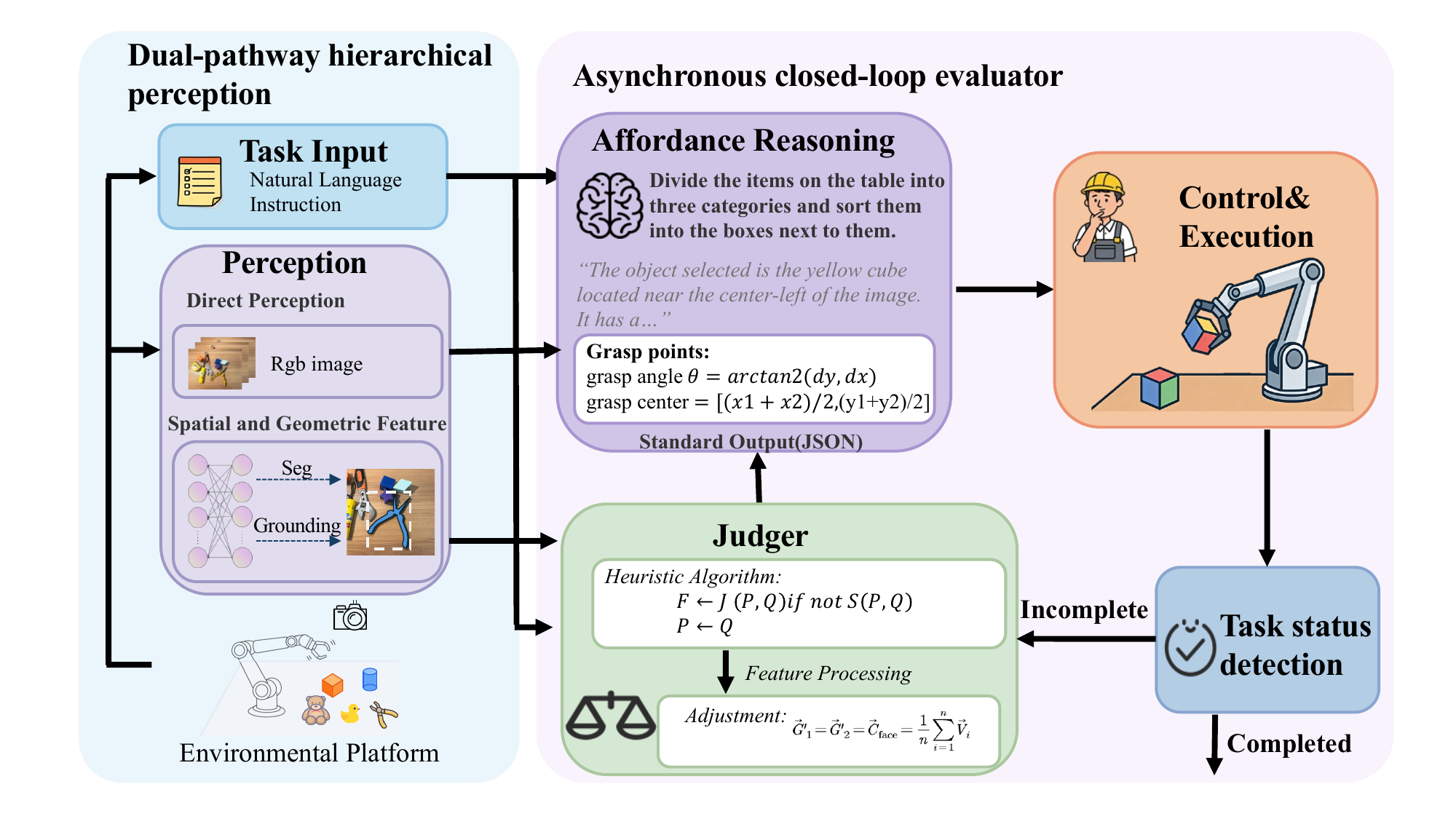}
  \caption{Overview of the functional components of the CLASP framework: The figure shows the core modules, from natural language instruction input and dual-path awareness to VLM-based heuristic reasoning and closed-loop feedback mechanism.}
  \label{fig:method_overview}
\end{figure*}

We define a closed-loop system $\mathcal{S}=(\mathcal{P},\mathcal{R},\mathcal{E})$ that maps an instruction $\ell$ and observation $I_t$ to an action $a_t$ and a discrete state $S_t$, i.e., $(a_t,S_t)=\mathcal{S}(\ell,I_t)$. The pipeline contains a Dual-Pathway Perception module $\mathcal{P}$, a Vision-Language Reasoner $\mathcal{R}$, and a Visual-Semantic Evaluator $\mathcal{E}$. The iteration continues until a terminal condition $S_t \in \mathcal{S}_{\text{term}}$ holds.

\textit{Definition 1} \textbf{Task-Oriented Grasp Objective}: Given a visual observation $I$, we define a task-conditioned energy functional $E(g;I)$ to evaluate any grasp $g \in \mathcal{G}$, where $\mathcal{G}$ represents the feasible grasp space. Task-oriented grasping refers to finding the optimal grasp $g^\star$ from this feasible set that minimizes the combined energy functional, namely,
\begin{equation}
g^\star = \arg\min_{g \in \mathcal{G}} E(g;I),
\end{equation}
subject to
\begin{equation}
E(g;I) = \lambda_g E_{geo}(g;I) + \lambda_s E_{sem}(g;I),
\end{equation}
where $E_{geo}(g;I)$ measures geometric grasp stability (e.g., antipodal alignment, local surface compatibility), and $E_{sem}(g;I)$ encodes functional affordance preferences (e.g., grasping the handle of a pan).

\textit{Definition 2} \textbf{Parallel-Jaw Grasp Parameterization}: To operationalize the grasp space $\mathcal{G}$, we assume a grasp executed by a parallel-jaw gripper is represented in the image space as two contact points $g = (p_1,p_2)$, where $p_1,p_2 \in \mathbb{R}^2$. The implied gripper width is $w(g) = \|p_1 - p_2\|$. A feasible grasp must satisfy the hardware's physical constraints, namely,
\begin{equation}
w_{min} \le w(g) \le w_{max},
\end{equation}
thus defining the feasible grasp set $\mathcal{G}$ as containing all grasps $g$ that satisfy these width boundaries.

\textbf{Scene and Observation Modeling}: We consider the task-oriented grasping of daily objects placed on a planar tabletop. The scene is defined in a world frame $\mathcal{F}_W=\{O_W,x_W,y_W,z_W\}$, with the table plane $\Pi=\{\mathbf{x}\in\mathbb{R}^3 \mid e_3^\top \mathbf{x}=0\}$, where $e_3=(0,0,1)^\top$. Stable placements restrict each object $\mathcal{O}_i$ to a 3-DoF manifold $\mathcal{C}_i=\{(x_i,y_i,\theta_i)\}$ (in-plane translation and yaw), while multi-object feasibility enforces non-penetration $d(\mathcal{O}_i,\mathcal{O}_j)\ge 0$. The observation $I$ (as formulated in Definition 1) is acquired via a calibrated top-down camera, with a small tilt modeled as $R_c=R_c^0\exp([\omega]_\times)$ and $\|\omega\|\le 10^\circ$.

\subsection{CLASP Design Overview}

Fig.~\ref{fig:method_overview} illustrates the architecture of the proposed framework for task-oriented robotic grasping. The pipeline initiates with natural language task inputs and a dual-pathway hierarchical perception module, which extracts direct RGB images alongside optional spatial and geometric features. These multimodal inputs are fed into the affordance reasoning module. Through semantic reasoning, this module generates a standardized JSON output containing precise grasp parameters, namely the grasp angle and center.

Subsequently, the predicted grasp parameters are sent to the control and execution module. After execution, an asynchronous closed-loop evaluator assesses the system state. If the object reaches the target category zone, the task terminates successfully. In the event of failure, the judger module is activated. It applies a heuristic procedure (Failure$\rightarrow$Shift$\rightarrow$Execute$\rightarrow$Check), together with feature processing, to compute spatial adjustments, such as moving grasp points toward the object's center. This diagnostic feedback is then routed back to the reasoning module, establishing a robust error-correction loop that dynamically improves subsequent attempts.
\begin{algorithm}[t]
\caption{CLASP}
\label{alg:simplified_pick_place}
\begin{algorithmic}[1]
\Require Simulator $sim$, max attempts $M$
\Ensure Success flag or count
\State $P \gets \varnothing,\ F \gets \emptyset$
\For{$a=1$ to $M$}
    \If{$P=\varnothing$}
        \State $(P,G)\gets \mathrm{DetectAndReason}{\mathrm{Capture}{sim}}$
    \Else
        \State $G\gets \mathrm{ReasonOnly}{P,F}$
    \EndIf
    \If{$P=\varnothing$} \State \textbf{break} \EndIf
    \State $g\gets \mathrm{ExtractGrasp}(P,G),\ x\gets \mathrm{Pix2Base}(g.center)$
    \State $t\gets \mathrm{GetPlaceTarget}(g.label)$; \If{$t=\varnothing$} \State \textbf{continue} \EndIf
    \If{\textbf{not} $\mathrm{ExecuteGrasp}(sim,x,g.angle,t)$} \State \textbf{continue} \EndIf
    \State $Q\gets \mathrm{DetectOnly}(\mathrm{Capture}(sim))$
    \If{$Q=\varnothing$ \textbf{or} $\mathrm{PlacementOK}(sim,t,g.label)$} \State \textbf{break} \EndIf
    \If{\textbf{not} $\mathrm{GraspSuccess}(P,Q)$}
        \State $F\gets \mathrm{JudgerFeedback}(P,Q)$
    \EndIf
    \State $P\gets Q$
\EndFor
\State \Call{ArmSleep}{sim}
\end{algorithmic}
\end{algorithm}

\subsubsection{Multimodal Dataset Construction.}
A data engine integrates large-scale real and synthetic desktop scenes to improve diversity and generalization. Each observation is mapped to $\mathcal{I}=(I_{\text{rgb}},D,\mathcal{B})$, where $I_{\text{rgb}}$ is the RGB image, $D$ is the depth map, and $\mathcal{B}$ is a sequence of semantic bounding boxes. For real-world scenes, $\mathcal{B}$ comes from semantic annotations and depth is fused as $D=\phi(D_{\text{sensor}},\hat{D}_{\text{rgb}})$, with $\hat{D}_{\text{rgb}}$ from RGB-based depth estimation. For synthetic scenes, $\mathcal{B}$ is extracted from rendered labels and $D=D_{\text{render}}$ is available directly. The Spatial Affordance Annotation module computes an interaction region set $\mathcal{R}=f(I_{\text{rgb}},D,\mathcal{B})$, yielding feasible graspable regions $\mathcal{R}_i$ for each object $\mathcal{O}_i$.

\subsubsection{Reasoning Template Annotation.}
We construct a labeled set $\mathcal{T}=\{(\tau_k,g_k,y_k)\}$ of Reasoning Template Annotations, where $\tau_k$ is the textual template, $g_k=\langle(x_1,y_1),(x_2,y_2)\rangle$ is the grasp point pair, and $y_k\in\{+1,-1\}$ indicates feasibility. Positive examples satisfy $y_k=+1$ with valid $g_k$, while negative examples satisfy $y_k=-1$ with invalid $g_k$ (e.g., NaN).

\subsubsection{Dual-Pathway Hierarchical Perception}
Let $I \in \mathbb{R}^{H \times W \times 3}$ be the RGB input. The perception module is a union of two mappings,
\begin{equation}
\mathcal{P}(I)=\mathcal{P}_{\text{sem}}(I)\ \cup\ \mathcal{P}_{\text{geo}}(I).
\end{equation}
The end-to-end semantic pathway produces a dense embedding $Z=\psi(I)$ for global inference. The geometric-aware pathway computes
\begin{equation}
\mathcal{B}=\mathrm{Det}(I,\ell),\quad \mathcal{M}=\mathrm{Seg}(I,\mathcal{B}),\quad \mathcal{G}=\mathrm{Geom}(\mathcal{M}),
\end{equation}
where $\mathcal{B}$ are open-vocabulary boxes, $\mathcal{M}$ are masks, and $\mathcal{G}$ are explicit pose/centroid descriptors, yielding reduced spatial uncertainty for downstream reasoning.

\subsubsection{VLM-Driven Structured Grasp Reasoning}
The reasoning module is a constrained predictor  fine-tuned on a large dataset.
\begin{equation}
A=\mathcal{R}(Z,\mathcal{G},\ell),\quad A=(u,v,\theta,c)\in\mathcal{A},
\end{equation}
where $(u,v)$ denote pixel coordinates, $\theta$ is the planar grasp angle, and $c$ is the inferred object category. We enforce a structured output space $\mathcal{A}$ via deterministic parsing (e.g., JSON) and incorporate spatial and collision constraints as predicates in $\mathcal{R}$.

\subsubsection{Closed-Loop Feedback and Execution}
The core of this part is a Judger that has been fine-tuned using an inference template based on the dataset.Given $A$, kinematic execution performs a projection $\pi:\mathbb{R}^2 \times \mathbb{R} \rightarrow \mathbb{R}^3$ using intrinsics and depth, i.e., $(X,Y,Z)=\pi(u,v,D)$, and the controller maps $(X,Y,Z,\theta)$ to a 7-DoF command $a_t$. After execution, the evaluator computes
\begin{equation}
\begin{aligned}
S_t &= \mathcal{E}(I_t,I_{t+1}) \\
&\in \{\text{Success}, \text{Adjustment Needed}, \text{Object Removed}\}
\end{aligned}
\end{equation}
and produces a textual rationale $\tau_t$ that is stored in the reasoner memory for the next iteration, forming an adaptive error-correction loop.
\section{Experimental Results}
\label{sec:experiments}

To evaluate the performance of various baseline models in executing pick-and-categorize tasks within cluttered scenarios, we conducted extensive evaluations. This section details the experimental setup, comparative results of the baselines, and a performance analysis under different constraints and environmental settings.

\begin{table}[htbp]
  \centering
  \caption{Simulation Environments and Evaluation Settings}
  \label{tab:env_settings}
  \setlength{\tabcolsep}{2pt}
  \begin{tabular}{@{}l >{\raggedright\arraybackslash}p{5.6cm}@{}}
    \hline
    \textbf{Field} & \textbf{Value} \\
    \hline
    Env. & Maniskill \\
    Robot & widowx \\
    Sim Freq & 500 \\
    Control Freq & 5 \\
    Static Friction & 0.5 \\
    Dynamic Friction & 0.5 \\
    Restitution & 0 (fixed at material creation) \\
    Damping & linear 0.1, angular 0.1 \\
    Pick Attempts & 3/2/1, 18/12/6 \\
    Pick Cls & 15 \\
    Target Cls & 3 \\
    Allowed Steps & Grasp-type: 80; PutOn/Stack-type: 120 \\
    Render Res & obs cam 128; render cam 512 \\
    View & third-person \\
    \hline
  \end{tabular}
\end{table}

\begin{figure}[htbp]
  \centering
  \includegraphics[width=\columnwidth]{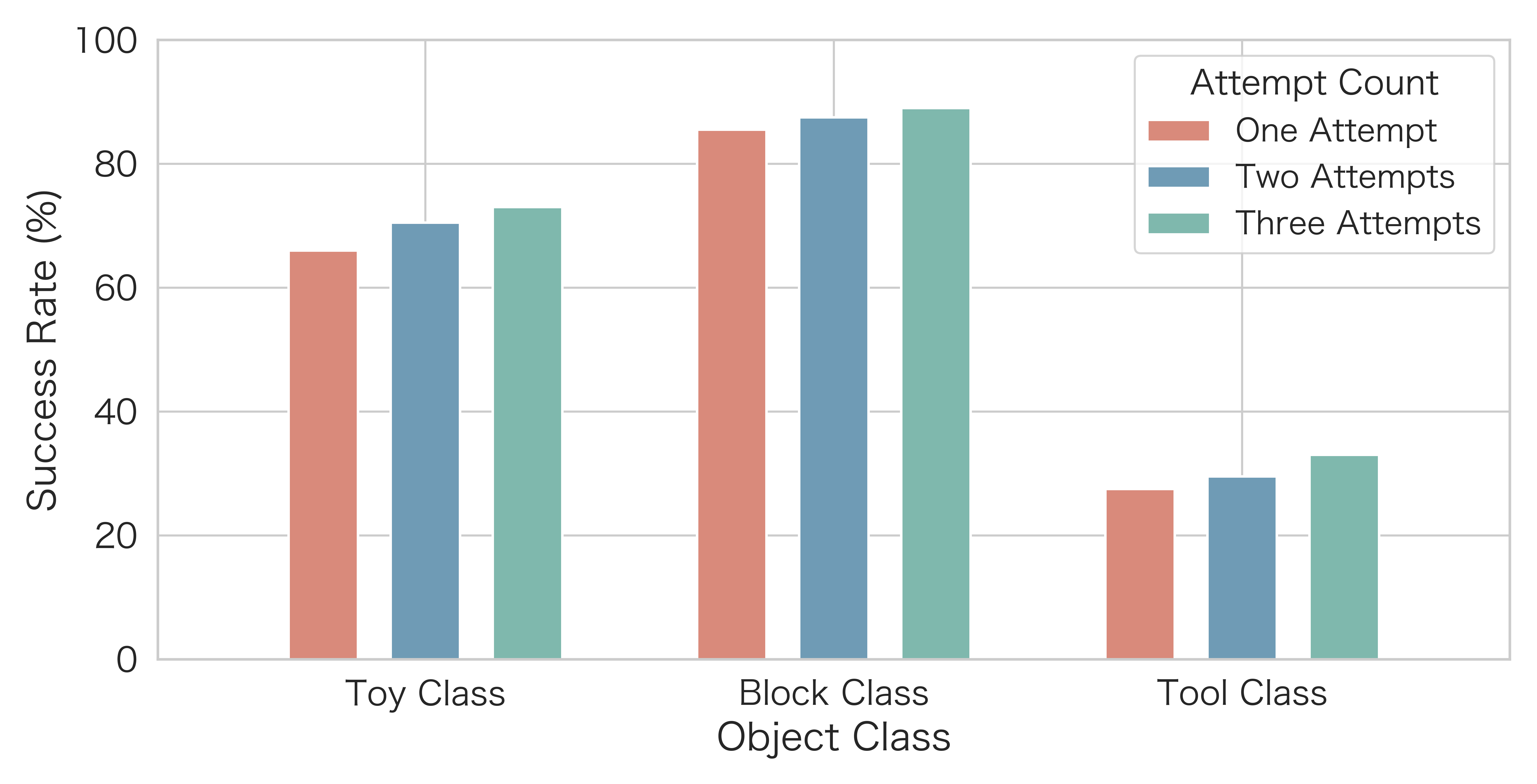}
  \caption{Pick success rates of different attempts.}
  \label{fig:Pick_success_rates}
\end{figure}

\begin{table*}[htbp]
  \centering
  \caption{Component Effectiveness in Pick Tasks}
  \label{tab:simpler_pick_all}
  \scriptsize 
  \setlength{\tabcolsep}{3pt}
  \begin{tabular}{lccccc|ccccc|cccccc}
    \hline
    \multirow{2}{*}{\textbf{Config}} & \multicolumn{5}{c}{\textbf{Toys}} & \multicolumn{5}{c}{\textbf{Blocks}} & \multicolumn{5}{c}{\textbf{Tools}} & \textbf{Total} \\
    \cline{2-6}\cline{7-11}\cline{12-16}
     & \textbf{banana} & \textbf{duck} & \textbf{lego} & \textbf{pear} & \textbf{teddy} & \textbf{cuboid} & \textbf{cylinder} & \textbf{semi-cylinder} & \textbf{ball} & \textbf{triangular} & \textbf{hammer} & \textbf{scissors} & \textbf{screwdriver} & \textbf{spatula} & \textbf{pliers} &  \\
    \hline
    R   & 10.0\% & \textbf{100.0\%} & \textbf{100.0\%} & 40.0\% & \textbf{100.0\%} & 20.0\% & 90.0\% & 90.0\% & \textbf{100.0\%} & \textbf{100.0\%} & 10.0\% & 20.0\% & 50.0\% & 0.0\% & 60.0\% & 59.3\% \\
    R+J  & 10.0\% & \textbf{100.0\%} & \textbf{100.0\%} & 40.0\% & 80.0\%  & 30.0\% & 80.0\% & 90.0\% & \textbf{100.0\%} & \textbf{100.0\%} & \textbf{40.0\%} & 30.0\% & 60.0\% & 0.0\% & 60.0\% & 61.3\% \\
    P+R+J & 20.0\% & \textbf{100.0\%} & \textbf{100.0\%} & 40.0\% & 90.0\%  & 80.0\% & \textbf{100.0\%} & \textbf{100.0\%} & \textbf{100.0\%} & \textbf{100.0\%} & 10.0\% & 0.0\% & 20.0\% & 20.0\% & 80.0\% & 64.0\% \\
    \textbf{P+R+H+J} & \textbf{60.0\%} & \textbf{100.0\%} & \textbf{100.0\%} & \textbf{80.0\%} & 90.0\%  & \textbf{100.0\%} & \textbf{100.0\%} & \textbf{100.0\%} & \textbf{100.0\%} & \textbf{100.0\%} & 20.0\% & \textbf{50.0\%} & \textbf{70.0\%} & \textbf{30.0\%} & \textbf{100.0\%} & \textbf{80.0\%} \\
    \hline
  \end{tabular}
  
  \vspace{4pt}
  \footnotesize \textit{Note:} R denotes Reasoner, J denotes Judger, H denotes Heuristic, and P denotes Perception Prior.
\end{table*}

\subsection{Data Sources}
The input combines two complementary datasets:
\begin{itemize}
  \item \textbf{200k+ real-world desktop scenes:} images collected from cluttered desktops, workbenches, and kitchen environments.
  \item \textbf{300k+ synthetic desktop scenes with 1000+ highprecision 3D models:} rendered scenes generated from a library of accurate object meshes, providing controllable appearance variations and clean geometry.
  \item \textbf{Dataset Overviews:} 

3D Object Interaction \cite{b41}: Comprising over 10,000 images and 50,000 annotations, this dataset predicts object mobility and interaction attributes to facilitate scene understanding.

EgoObjects \cite{b42}: A massive first-person benchmark containing over 9,000 videos and 650,000 annotations for real-world object recognition and segmentation amidst natural clutter.

GraspNet \cite{b38}: Featuring nearly 100,000 RGB-D images and nearly 370 million grasp poses, this is a foundational resource for training robotic grasp planning algorithms.

Open X-Embodiment \cite{b9}: Integrating over 1 million real-world trajectories across 22 diverse robot platforms, it enables the training of generalized, cross-morphology control policies.

RLBench \cite{b43}: A simulation framework providing over 100 diverse manipulation tasks with massive procedural demonstration generation to train and evaluate robot learning algorithms.
\end{itemize}

\subsection{Experimental Setup and Metrics}
\textbf{Simulation Environments:} We conducted all experiments in Maniskill. The simulator uses a WidowX robot with a 500 Hz simulation frequency and a 5 Hz control frequency. The environment settings include static and dynamic friction of 0.5, zero restitution, and linear/angular damping of 0.1. We allowed up to 15 pick classes and 3 target classes; grasp-type and put-on/stack-type episodes permit 80 and 120 steps, respectively. Notably, all 15 evaluation objects are unseen with respect to the training dataset. The camera settings use observation resolution 128 and render resolution 512 with a third-person view. The simulation environment configuration is shown in Table~\ref{tab:env_settings}.

\textbf{Task Definition and Evaluation Metrics:} We define three high-level categories---Toys, Tools, and Blocks---and use 15 common everyday objects with distinct geometric properties as targets for both grasping and categorization evaluation. Each trial starts from a cluttered scene; the robot must pick the specified object and place it into its corresponding category zone. A task is counted as successful only if the object is grasped without being dropped and is placed in the correct zone. We employ two core metrics for quantitative evaluation:

1) Pick Success Rate.
This metric measures the robot's basic manipulation reliability, defined as the proportion of successful grasp-and-lift operations:
\begin{equation}
\text{Pick Success Rate} = \frac{N_{\text{succ}}}{N_{\text{att}}},
\end{equation}
where $N_{\text{succ}}$ denotes the number of successful grasps and $N_{\text{att}}$ is the total number of grasp attempts.

2) Mean Intersection over Union (mIoU).
This metric evaluates placement precision and semantic correctness by measuring the spatial overlap between the placed object region and the designated category zone:
\begin{equation}
\text{mIoU} =
\frac{\lvert \mathcal{A}_{\text{obj}} \cap \mathcal{A}_{\text{zone}} \rvert}
{\lvert \mathcal{A}_{\text{obj}} \cup \mathcal{A}_{\text{zone}} \rvert},
\end{equation}
where $\mathcal{A}_{\text{obj}}$ represents the region of the placed object and $\mathcal{A}_{\text{zone}}$ denotes the corresponding target zone.

\textbf{Baseline:} 
For fairness, we note that the VLM baselines are trained on a large-scale multimodal corpus containing availability-related categories and annotations; therefore, it is fair to compare the two VLMs in the baselines below with our method.

\textbf{Qwen3-VL}\cite{b26} is an open-source, large-scale visual language model from Alibaba Cloud, which supports image understanding, visual reasoning, text recognition, etc.
\textbf{Doubao-Seed}\cite{b40} is a self-developed large multimodal foundation model by ByteDance, featuring strong multimodal understanding, long-context reasoning, etc.
\textbf{AffordanceNet}\cite{b39} is an end-to-end deep learning framework that simultaneously detects objects and predicts their affordance masks from RGB images.
\textbf{Anygrasp}\cite{b10} is an open-source method for 3D robotic grasping that predicts grasp poses directly from point clouds.

While the \textbf{Visual-Language-Action(VLA)} models also perform well in grasping desktop objects, they are highly dependent on post-training. According to our research, no VLA model currently can achieve zero-shot generalization to grasping common desktop objects with only pre-training. Therefore, considering the fairness of the grasping comparison experiment, we did not choose to fine-tune the existing VLA for baseline comparison.

\subsection{Performance of General Pick Task}
We compared the pick success rates using objects including banana, duck, lego, pear, and teddy. In Table~\ref{tab:simpler_pick_all}, R denotes Reasoner, J denotes Judger, H denotes Heuristic, and P denotes Perception Prior. Importantly, all results in Table~\ref{tab:simpler_pick_all} are obtained by directly testing off-the-shelf models without additional task-specific fine-tuning. The table illustrates the performance of various models under different maximum attempt constraints. Some of these toys and blocks have relatively symmetrical geometric features. A small rotation of the pick angle does not result in failure, so the pick success rate is relatively high, reaching 100\%. Objects that have strict position requirements for the pick operation or are relatively heavy, such as shovels, scissors, and hammers, have a relatively low pick success rate. Therefore, although our method improves the success rate of tool picking compared to other baselines, the score is still relatively low.

\paragraph{The Impact of Different Pickup Objects}

Fig.~\ref{fig:Pick_success_rates} summarizes the average pick success rates over three object classes (Toy, Block, and Tool) under different maximum attempt trials ($\{1,2,3\}$). Across all attempt settings, Block objects remain the easiest category (85.5\%/87.5\%/89.0\%), followed by Toy objects (66.0\%/70.5\%/73.0\%), while Tool objects are consistently the most challenging (27.5\%/29.5\%/33.0\%).
Increasing the attempt budget from 1 to 3 improves performance for all classes, with gains of +7.0\% (Toy), +3.5\% (Block), and +5.5\% (Tool), and raises the class-averaged success rate from 59.7\% to 65.0\%.
These results indicate that a larger attempt budget provides a stable but limited improvement, and that category-specific difficulty—particularly for tools—remains the dominant factor governing task success.

\paragraph{The Success Rate of Category-level Pickup}
Table~\ref{tab:simpler_pick_all} reports category-level pick success rates across Toys, Blocks, and Tools. The full pipeline (P+R+H+J) consistently achieves the strongest performance, especially on geometrically challenging Tool objects (e.g., hammer/scissors/screwdriver/spatula/pliers) where partial pipelines show larger drops. Adding Perception Prior (P) improves stability for block-like shapes (cuboid/cylinder/semi-cylinder/ball/triangular), while the Heuristic module (H) further boosts overall robustness, yielding the highest total score among all configurations. These trends indicate that perceptual priors and heuristic reasoning jointly reduce failure cases in cluttered scenes and improve cross-category generalization.

\paragraph{The Success Rate of Method-level Pickup}

\begin{table}[htbp]
\centering
\caption{Method Performance Comparison in General Pick Tasks}
\label{tab:method_comparison}
\begin{tabular}{lcccc}
\hline
Method & Toys & Blocks & Tools & Overall \\
\hline
Qwen3-VL~\cite{b26} & 92.00\%  & 78.40\% & 46.40\% & 72.27\% \\
Doubao-Seed~\cite{b40} & 88.00\% & 66.00\% & 52.00\% & 68.67\% \\
AffordanceNet\cite{b39} & \textbf{100.00\%} & 72.00\% & 54.00\% & 75.33\% \\
AnyGrasp~\cite{b10} & 63.20\%  & 62.40\% & 44.00\% & 56.53\% \\
CLASP(w/o pretrain) & 86.00\%  & \textbf{100.00\%} & 54.00\% & 80.00\% \\
\textbf{CLASP(Ours)} & \textbf{100.00\%} & {84.00\%} & \textbf{77.00\%} & \textbf{87.00\%} \\
\hline
\end{tabular}
\end{table}

In the comparative experiments, we input natural language prompts and RGB images into Qwen3-VL, Doubao-Seed, and AffordanceNet. Qwen3-VL and Doubao-Seed directly output end-to-end grasp points, while AffordanceNet predicts affordance masks that are aligned with the robotic arm's 7-DoF actions through its action module. AnyGrasp does not include semantic understanding; therefore, we only input scene images and 3D features into AnyGrasp and align the predicted grasps with the robotic arm's 7-DoF actions. Table~\ref{tab:method_comparison} presents a method-level comparison for the general pick task. The CLASP (w/o pretrain) results already demonstrate the effectiveness of our framework design. After that, we fine-tune the model on our own dataset and obtain the best overall performance (87.00\%). The above experiments fully demonstrate the effectiveness of the CLASP architecture design and data pre-training, as well as its good generalization performance.
Notably, the Blocks score decreases after fine-tuning(from 100.00\% to 84.00\%), mainly because our training dataset contains relatively limited affordance data for block objects, which weakens category-specific adaptation in that class.It is also noteworthy that all methods exhibited the lowest success rate for retrieving Tools. This is because tools themselves have more complex geometry compared to other everyday objects, requiring more robust semantic information for retrieval, thus increasing the difficulty. Our solution, however, achieved the highest success rate (77\%) on tools, demonstrating the high gain effect of the solution in utilizing semantic information.

\subsection{Simulated System Latency Benchmark}
To assess latency, we implemented two execution policies: Streaming and Asynchronous. The \textbf{Streaming Policy} employs single-threaded sequential execution with blocking steps. In contrast, the \textbf{Asynchronous Policy} we proposed decouples Reasoner and Judger modules into independent background threads, enabling asynchronous perception and reasoning during execution, thereby minimizing blocking overhead. The evaluation was conducted on 15 objects with diverse characteristics under a standardized pick-and-place task. Latency performance was measured using three metrics: total execution time, average execution time, and variance.

As shown in Table~\ref{tab:framework_comparison}, compared with the Streaming Policy, the Asynchronous Policy achieves a 39.8\% reduction in total execution time and a 39.9\% reduction in average task completion time, demonstrating a substantial improvement in execution efficiency. In addition, execution-time variance is reduced by 59.5\%, indicating markedly lower temporal fluctuation and improved operational stability under the asynchronous architecture. The results strongly demonstrate the effectiveness of our asynchronous strategy in improving the efficiency of action execution and interaction.

\begin{table}[htbp]
  \centering
  \caption{Latency Benchmark: Streaming vs. Asynchronous Execution}
  \label{tab:framework_comparison}
  \begin{tabular}{lccc}
    \hline
    \textbf{Policy} & \textbf{Exec. Time (s)} & \textbf{Avg. Time (s)} & \textbf{Var. (s$^2$)} \\
    \hline
    Streaming Policy & 1613.3 & 107.6 & 1353.3 \\
    \textbf{Asynchronous Policy} & \textbf{970.9} & \textbf{64.7} & \textbf{547.8} \\
    \hline
  \end{tabular}
\end{table}

\subsection{Categorization and Overall Task Success Rate}

\begin{table}[htbp]
  \centering
  \caption{Performance comparison on different object categories}
  \label{tab:success_rate}
  \begin{tabular}{l c c c c}
    \hline
    Method & Toys & Tools & Blocks & Total \\
    \hline
    Qwen3-VL~\cite{b26} & 75.00\% & 50.00\% & 60.00\% & 63.33\% \\
    Doubao-Seed~\cite{b40} & 48.33\% & 58.33\% & 30.00\% & 46.67\% \\
    \textbf{CLASP(Ours)} & \textbf{85.00\%} & \textbf{70.00\%} & \textbf{86.67\%} & \textbf{83.33\%} \\
    \hline
  \end{tabular}
\end{table}

\begin{figure}[ht]
  \centering
  \includegraphics[width=\columnwidth]{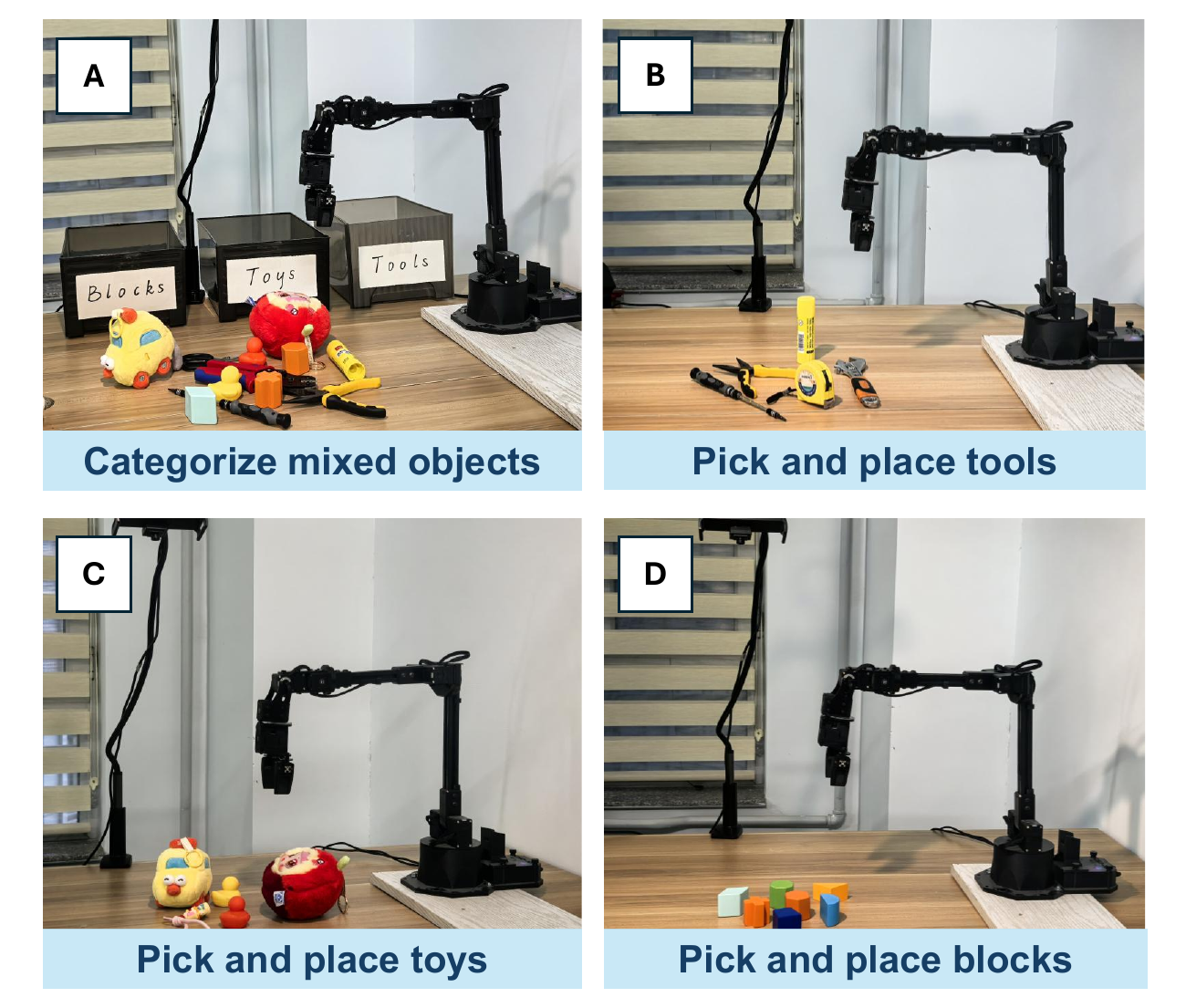}
  \caption{The WidowX 250S 6-DoF robotic manipulator used in real-world experiments. (A) shows the performance of WidowX 250S 6-DoF robotic manipulator categorize mixed objects, selecting and placing tools(B), toys(C), and blocks(D) in a real environment}
  \label{fig:widowx_setup}
\end{figure}

\begin{figure}[htbp]
  \centering
  \includegraphics[width=\columnwidth]{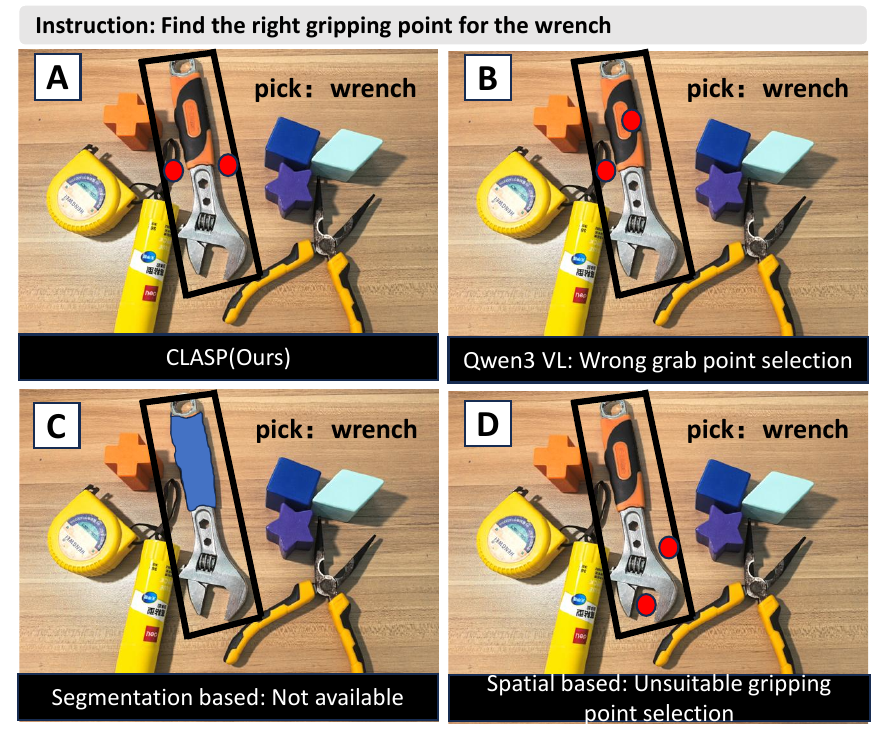}
  \caption{Our CLASP method demonstrates superior effectiveness in physical settings, particularly in the precision and suitability of grasp point selection. While baseline methods show clear deficiencies—such as (B)Qwen3-VL selecting an incorrect grasp point, (C)segmentation-based methods failing to produce an available result, and (D)spatial-based methods identifying unsuitable, non-graspable positions on the edge—our method accurately determines the optimal opposing grasp locations on the wrench handle.}
  \label{fig:gripping_point}
\end{figure}

The categorization task evaluates both correct class selection and accurate placement. Table~\ref{tab:success_rate} indicates that the classification IoU of our model is improved by 10.0\%, 20.0\% and 26.67\% respectively compared with the single Qwen3-VL model, and the overall success rate is increased by 20.00\%. Overall, these results confirm that our pipeline improves categorical placement reliability and reduces sensitivity to geometric variation across object types.

\subsection{Real World Implementation}

All experiments were conducted on a WidowX 250S 6-DoF robotic manipulator with low-level motion control implemented via ROS, as illustrated in Fig.~\ref{fig:widowx_setup}. Visual observations were captured by a USB RGB camera mounted above the workspace, providing a stable top-down view. The task scenarios encompass three categories of objects with distinct geometric features and textures: Toys, Tools, and Blocks, which correspond to the dataset used in our simulation evaluation task.

As shown in Fig.~\ref{fig:gripping_point}, real-world validation provides compelling evidence that our method achieves superior effectiveness in physical settings, particularly in the precision and suitability of grasp point selection. The baseline methods show clear deficiencies in complex scenarios: Qwen3 VL (B) selects a wrong grasp point; the segmentation-based method (C) cannot produce an available result; and the spatial-based method (D) identifies an unsuitable and non-graspable position on the edge. Conversely, our CLASP method (A) accurately determines opposing and optimal grasp locations on the wrench handle. These results conclusively demonstrate that our system not only mitigates common pitfalls such as misgrasps and target confusion but also establishes a reliable, stable, and robust deployability on real hardware platforms.
\section*{Conclusion}

In this paper, we presented a novel closed-loop framework to address the critical bottlenecks of deploying VLMs in low-level robotic manipulation, specifically spatial hallucinations and the fragility of open-loop execution. First, we introduced a scalable multi-modal data engine that automatically synthesizes high-quality demonstration data without requiring human teleoperation. Second, we designed a Dual-Pathway Hierarchical Perception module that successfully decouples high-level semantic intent from fine-grained geometric grounding. Finally, we implemented an Asynchronous Closed-Loop Evaluator to compare pre- and post-execution states, providing text-based diagnostic feedback to establish a robust error-correction mechanism. Extensive experiments in the ManiSkill simulator demonstrate that our approach achieves an 87.0\% overall success rate. Notably, the proposed framework exhibits remarkable generalization across diverse objects, providing exceptional robustness in geometrically challenging categories , specifically tool-like objects. Future work will explore deploying the framework on physical robots to further validate its real-world applicability and extending the reasoning capabilities to longer-horizon manipulation tasks.

\bibliographystyle{IEEEtran}
\bibliography{references}
\vspace{12pt}
\color{red}

\end{document}